\documentclass{article}
\usepackage{hyperref}
\usepackage{url}
\usepackage{graphicx,color}
\usepackage{paralist}
\usepackage{amsmath}
\usepackage[letterpaper, left=1cm, right=1.25cm, top=1cm, bottom=1.25cm]{geometry}
\usepackage{setspace}
\usepackage{subfigure}
\usepackage{fancyhdr}

\title{Lesion Border Detection in Dermoscopy Images}
\author{%
        M. Emre Celebi\\
        Dept. of Computer Science\\Louisiana State Univ., Shreveport, LA, USA\\
        \href{mailto:ecelebi@lsus.edu}{ecelebi@lsus.edu}
\and
        Hitoshi Iyatomi\\
        Dept. of Electrical Informatics\\Hosei Univ., Tokyo, Japan\\
        \href{mailto:iyatomi@hosei.ac.jp}{iyatomi@hosei.ac.jp}
\and
        Gerald Schaefer\\
        School of Engineering and Applied Science\\Aston Univ., Birmingham, UK\\
        \href{mailto:g.schaefer@aston.ac.uk}{g.schaefer@aston.ac.uk}
\and
        William V. Stoecker\\
        Stoecker \& Associates, Rolla, MO, USA\\
        \href{mailto:wvs@mst.edu}{wvs@mst.edu}
       }
\begin{document}
\maketitle
%\linenumbers
\begin{abstract}

\textbf{Background:} Dermoscopy is one of the major imaging modalities used in the diagnosis of melanoma and other pigmented skin lesions. Due to the difficulty and subjectivity of human interpretation, computerized analysis of dermoscopy images has become an important research area. One of the most important steps in dermoscopy image analysis is the automated detection of lesion borders. \textbf{Methods:} In this article, we present a systematic overview of the recent border detection methods in the literature paying particular attention to computational issues and evaluation aspects. \textbf{Conclusion:} Common problems with the existing approaches include the acquisition, size, and diagnostic distribution of the test image set, the evaluation of the results, and the inadequate description of the employed methods. Border determination by dermatologists appears to depend upon higher-level knowledge, therefore it is likely that the incorporation of domain knowledge in automated methods will enable them to perform better, especially in sets of images with a variety of diagnoses.
\end{abstract}
%\pagebreak[4]

\section{Introduction}
\label{sec_intro}
Invasive and in-situ malignant melanoma together comprise one of the most rapidly increasing cancers in the world. Invasive melanoma alone has an estimated incidence of 62,480 and an estimated total of 8,420 deaths in the United States in 2008~\cite{Jemal07}. Early diagnosis is particularly important since melanoma can be cured with a simple excision if detected early.
\par
Dermoscopy, also known as epiluminescence microscopy, is a non-invasive skin imaging technique that uses optical magnification and either liquid immersion and low angle-of-incidence lighting or cross-polarized lighting, making subsurface structures more easily visible when compared to conventional clinical images~\cite{Argenziano02}. Dermoscopy allows the identification of dozens of morphological features such as pigment networks, dots/globules, streaks, blue-white areas, and blotches~\cite{Menzies03}. This reduces screening errors, and provides greater differentiation between difficult lesions such as pigmented Spitz nevi and small, clinically equivocal lesions~\cite{Steiner93}. However, it has been demonstrated that dermoscopy may actually lower the diagnostic accuracy in the hands of inexperienced dermatologists~\cite{Binder95}. Therefore, in order to minimize the diagnostic errors that result from the difficulty and subjectivity of visual interpretation, the development of computerized image analysis techniques is of paramount importance~\cite{Fleming98}.
\par
\begin{figure}[!ht]
\centering
 \subfigure[Low contrast]{\label{fig_problems_a}\includegraphics[width=0.2\columnwidth]{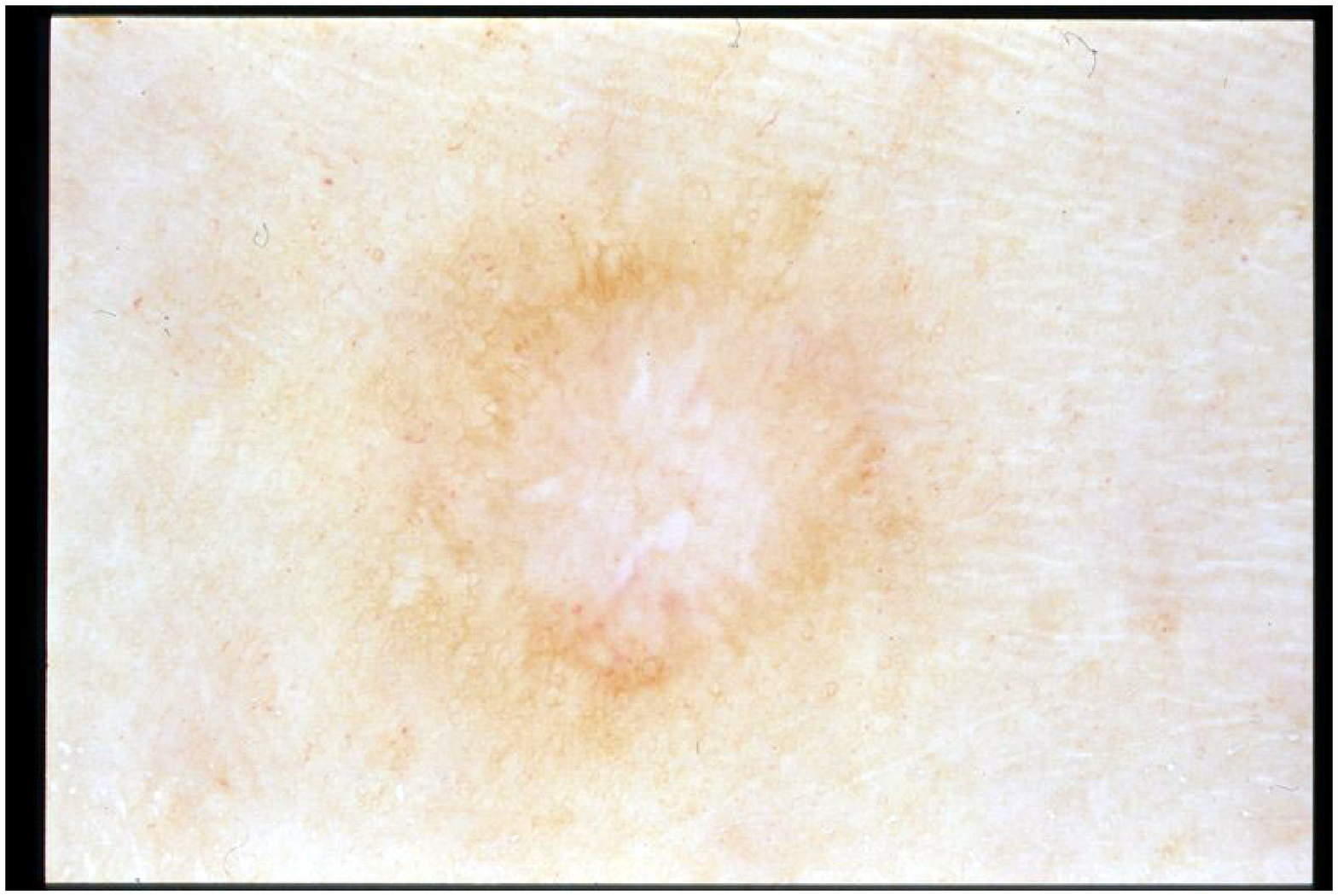}}
 \hspace{.1in}
 \subfigure[Irregular border]{\label{fig_problems_b}\includegraphics[width=0.2\columnwidth]{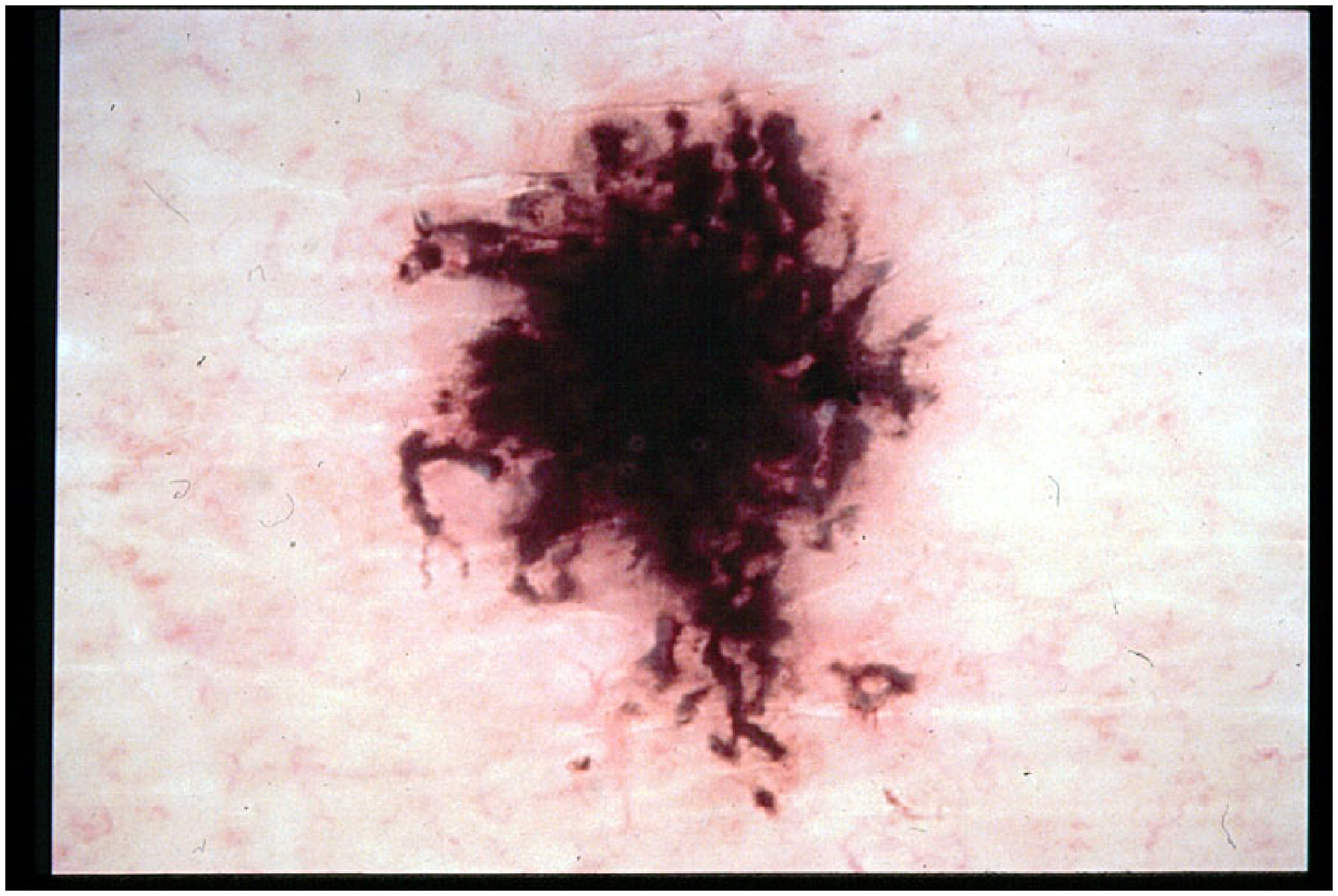}}
 \hspace{.1in}
 \subfigure[Fuzzy border]{\label{fig_problems_c}\includegraphics[width=0.2\columnwidth]{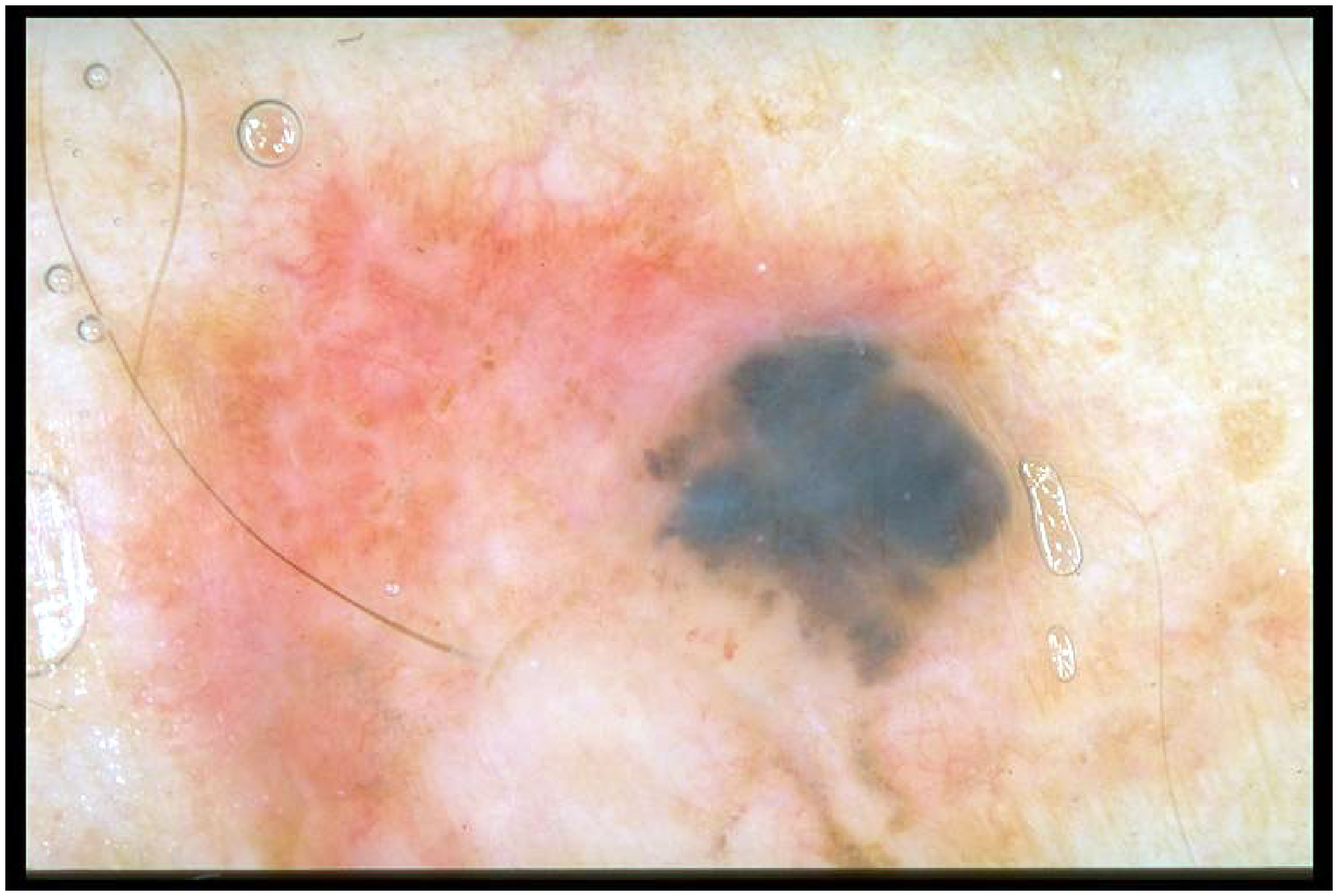}}
 \hspace{.1in}
 \subfigure[Blood vessels]{\label{fig_problems_d}\includegraphics[width=0.2\columnwidth]{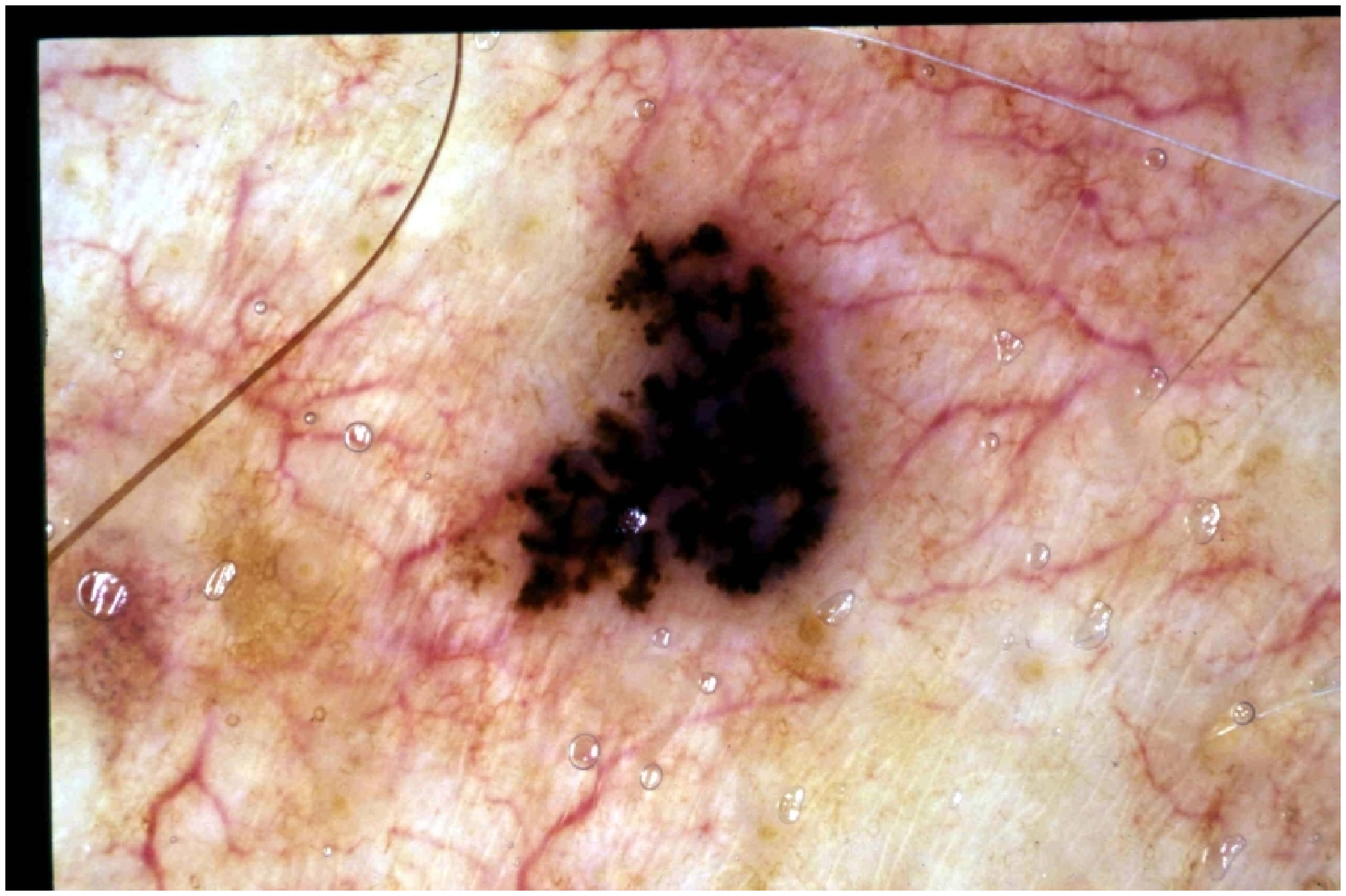}}
 \hspace{.1in}
 \subfigure[Hairs]{\label{fig_problems_e}\includegraphics[width=0.2\columnwidth]{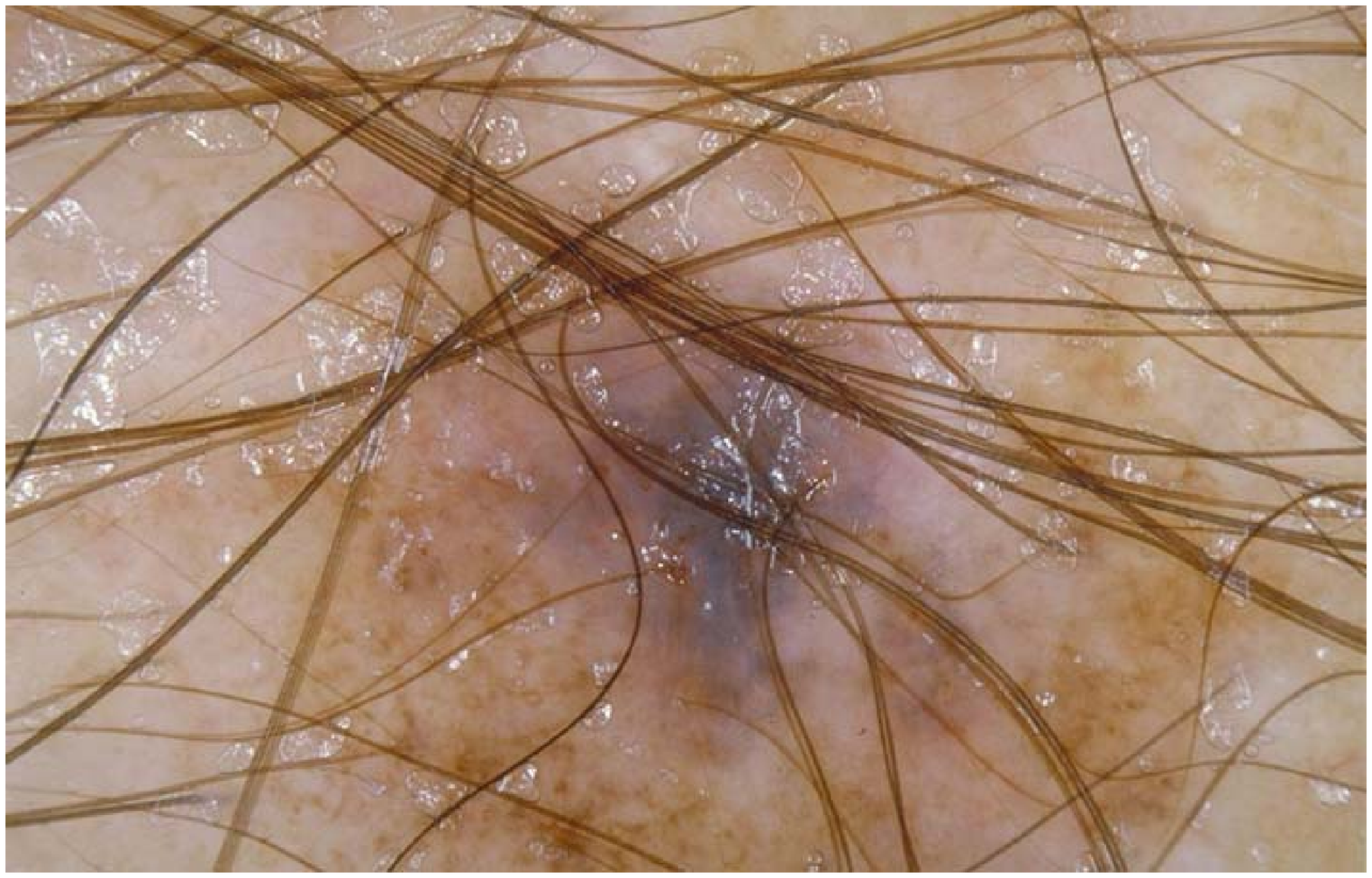}}
 \hspace{.1in}
 \subfigure[Bubbles]{\label{fig_problems_f}\includegraphics[width=0.2\columnwidth]{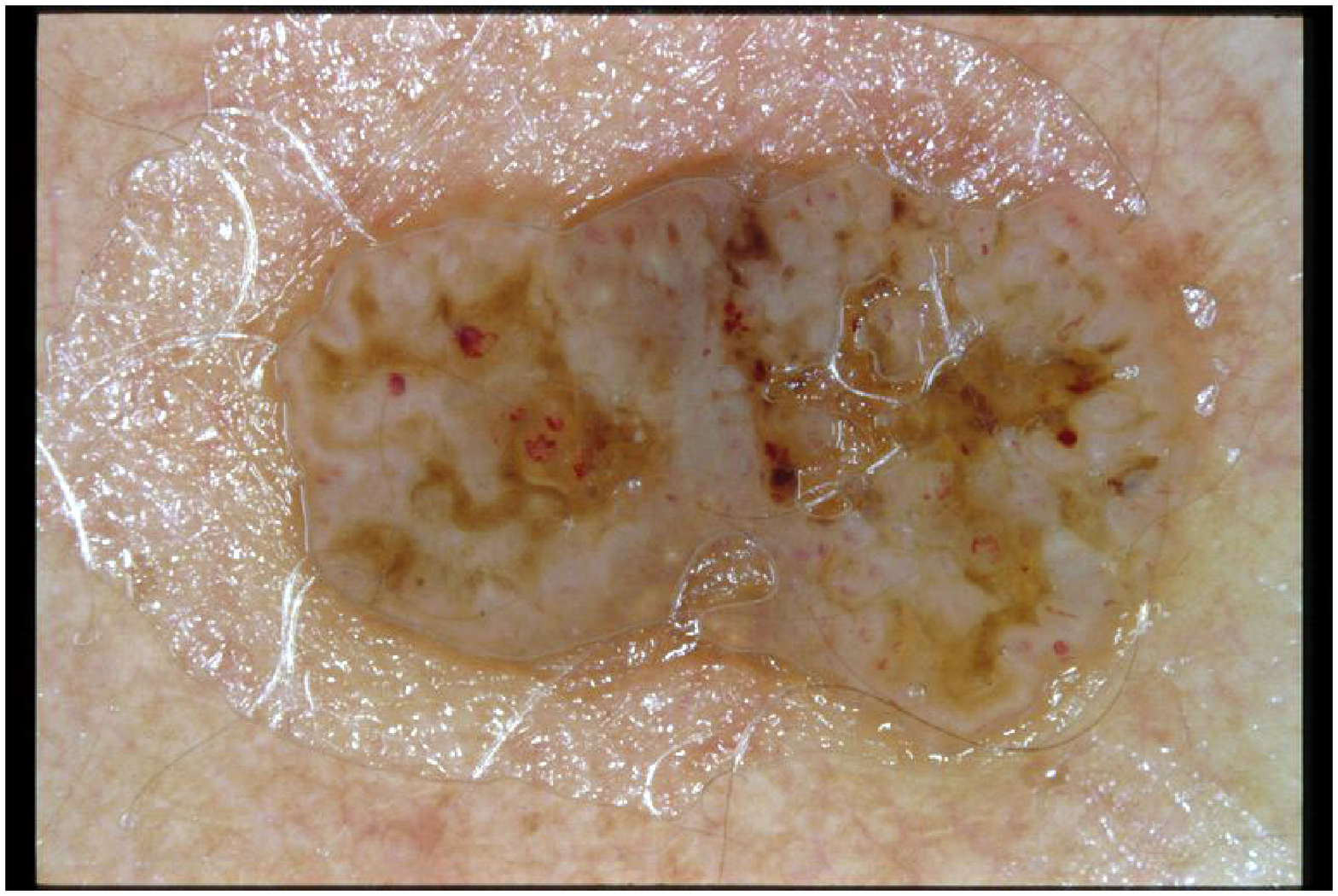}}
 \hspace{.1in}
 \subfigure[Variegated coloring]{\label{fig_problems_g}\includegraphics[width=0.2\columnwidth]{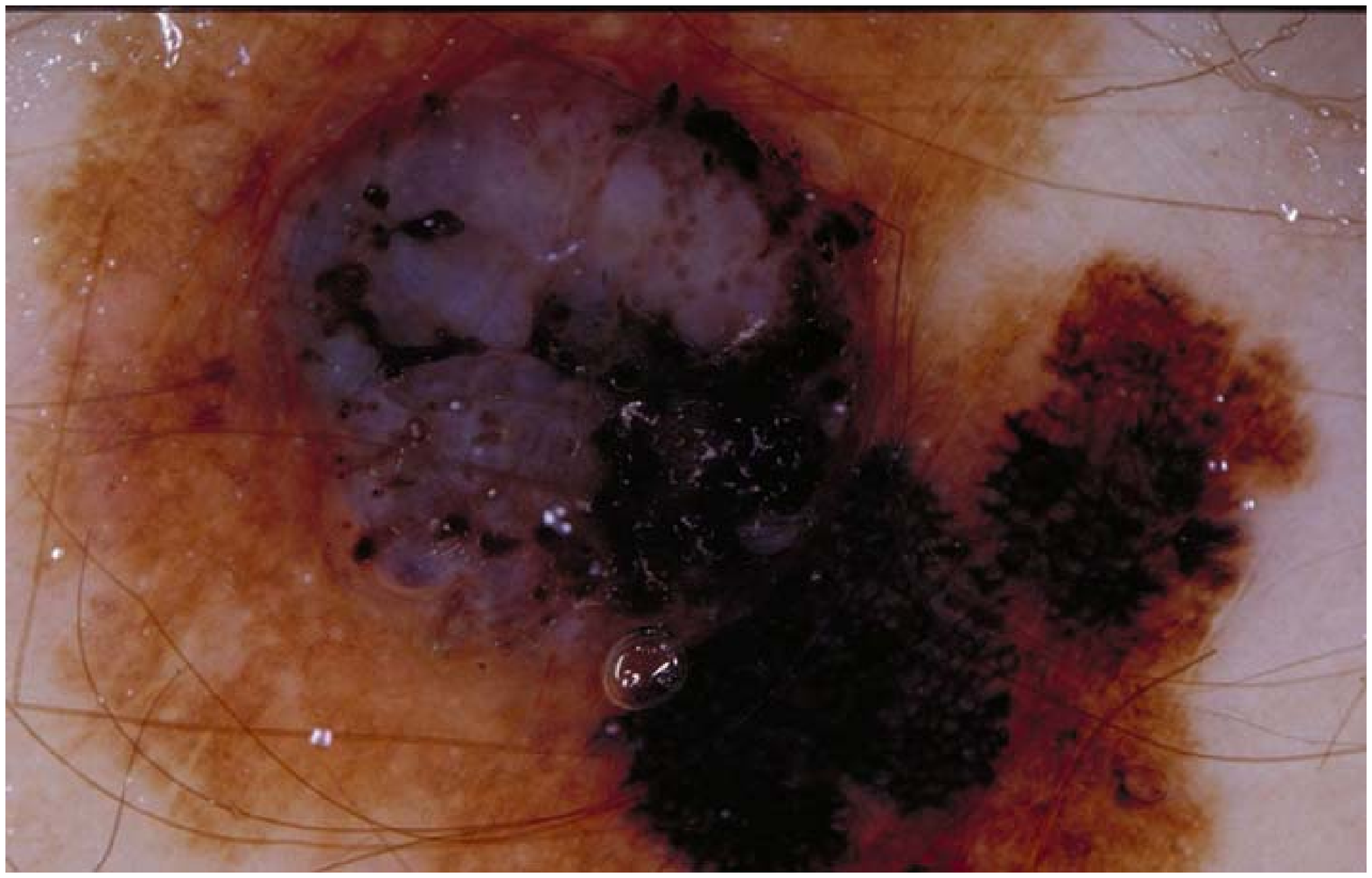}}
 \hspace{.1in}
 \subfigure[Fragmentation]{\label{fig_problems_h}\includegraphics[width=0.2\columnwidth]{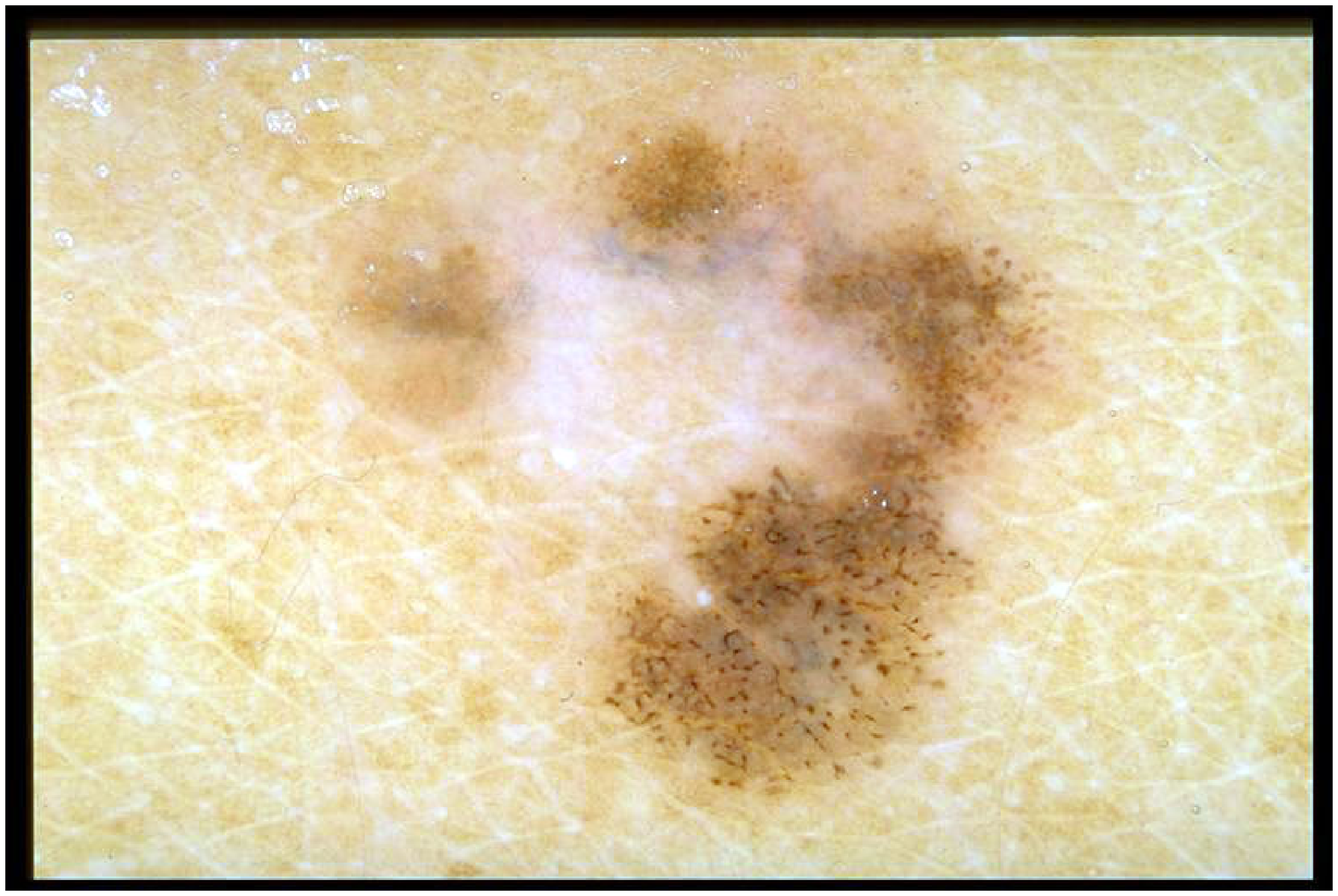}}
 \caption{Problems with border detection}
 \label{fig_problems}
\end{figure}

Automated border detection is often the first step in the automated or semi-automated analysis of dermoscopy images~\cite{Celebi07a}. It is crucial for the image analysis for two main reasons. First, the border structure provides important information for accurate diagnosis, as many clinical features such as asymmetry, border irregularity, and abrupt border cutoff are calculated directly from the border. Second, the extraction of other important clinical features such as an atypical pigment network, globules, and blue-white areas, critically depends on the accuracy of border detection. Automated border detection is a challenging task due to several reasons:
\begin{inparaenum}[(i)]
\item low contrast between the lesion and the surrounding skin (Fig.~\ref{fig_problems_a}),
\item irregular (Fig.~\ref{fig_problems_b}) and fuzzy lesion borders (Fig.~\ref{fig_problems_c}),
\item artifacts and intrinsic cutaneous features such as black frames, skin lines, blood vessels (Fig.~\ref{fig_problems_d}), hairs (Fig.~\ref{fig_problems_e}), and air bubbles (Fig.~\ref{fig_problems_f}),
\item variegated coloring inside the lesion (Fig.~\ref{fig_problems_g}), and
\item fragmentation due to various reasons such as scar-like depigmentation (Fig.~\ref{fig_problems_h}).
\end{inparaenum}

In the last decade, numerous methods have been developed for automated border detection in dermoscopy images. In this article, we present an overview of recent border detection methods and describe the preprocessing, segmentation, and postprocessing steps involved in each method. We also review performance evaluation issues and propose guidelines for studies in automated border detection.
\section{Preprocessing}
\label{sec_pre}
In this section, we describe the preprocessing steps that facilitate the border detection procedure namely, color space transformation, contrast enhancement, and artifact removal.
\subsection{Color Space Transformation}
Dermoscopy images are commonly acquired using a digital camera with a dermoscope attachment. Due to the computational simplicity and convenience of scalar (single channel) processing, the resulting RGB (red-green-blue) color image is often converted to a scalar image using one of the following methods:
\begin{itemize}
	\item Retaining only the blue channel (lesions are often more prominent in this channel).
	\item Applying the luminance transformation, i.e. $\mbox{Luminance} = 0.299 \times \mbox{Red} + 0.587 \times \mbox{Green} + 0.114 \times \mbox{Blue}$.
	\item Applying the Karhunen-Lo\'{e}ve (KL) transformation~\cite{Pratt07} and retaining the channel with the highest variance.
\end{itemize}
In applications where vector (multichannel) processing is desired, the RGB image can be used directly or it might be transformed into a different color space for various reasons including:
\begin{inparaenum}[(i)]
 \item reducing the number of channels,
 \item decoupling luminance and chromaticity information,
 \item ensuring approximate perceptual uniformity, and
 \item achieving invariance to different imaging conditions such as viewing direction, illumination intensity, and highlights.
\end{inparaenum}
Common target color spaces in this case include CIEL*a*b*, CIEL*u*v*, KL, and HSI (Hue-Saturation-Intensity)~\cite{Pratt07}. Note that the use of the CIEL*a*b* and CIEL*u*v* color spaces requires careful calibration of the acquisition device, a step that seems to be frequently neglected in the literature.
\subsection{Contrast Enhancement}
As mentioned in Section~\ref{sec_intro}, one of the factors that complicate the detection of borders in dermoscopy images is insufficient contrast. Recently, Delgado \emph{et al.}~\cite{Delgado08} proposed a contrast enhancement method based on independent histogram pursuit (IHP). This algorithm linearly transforms the original RGB image to a decorrelated color space in which the lesion and the background skin are maximally separated. Border detection is then performed on these contrast-enhanced images using a simple clustering algorithm.
\subsection{Artifact Removal}
Dermoscopy images often contain artifacts such as such as black frames, ink markings, rulers, air bubbles, as well as intrinsic cutaneous features that can affect border detection such as blood vessels, hairs, and skin lines. These  artifacts and extraneous elements complicate the border detection procedure, which results in loss of accuracy as well as an increase in computational time. The most straightforward way to remove these artifacts is to smooth the image using a general purpose filter such as the Gaussian (GF), median (MF), or anisotropic diffusion filters (ADF). Several issues should be considered while using these filters:
\begin{itemize}
	\item Scalar vs.\ vector processing: These filters are originally formulated for scalar images. For vector images one can apply a scalar filter to each channel independently and then combine the results, a strategy referred to as \emph{marginal} filtering. Although fast, this scheme introduces color artifacts in the output. An alternative solution is to use filters that treat the pixels as vectors~\cite{Celebi07c}.
	\item Mask size: The amount of smoothing is proportional to the mask size. However, excessively large masks result in the blurring of edges, which might reduce the border detection accuracy. Setting the mask size proportional to the image size seems to be a reasonable strategy~\cite{Schmid99a,Celebi08a}.
	\item Computational time: For the GF and MF, algorithms that perform in constant time independent of the mask size have been developed~\cite{Geusebroek03,Perreault07}. As for the ADF, the computational time depends on the mask size and the number of iterations.
\end{itemize}
An alternative strategy for artifact removal is to use specialized methods for each artifact type. For the removal of black frames, Celebi \emph{et al.}~\cite{Celebi08a} proposed an iterative algorithm based on the lightness component of the HSL (Hue-Saturation-Lightness) color space. In most cases, image smoothing  effectively removes the skin lines and blood vessels. Hair removal received the most attention in the literature. Lee \emph{et al.}~\cite{Lee97} and Schmid~\cite{Schmid99a} used mathematical morphology. Fleming \emph{et al.}~\cite{Fleming98} applied curvilinear structure detection with various constraints followed by gap filling. Recently, Zhou \emph{et al.}~\cite{Zhou08} and Wighton \emph{et al.}~\cite{Wighton08} proposed more sophisticated approaches based on inpainting. A method that can remove bubbles with bright edges was introduced in~\cite{Fleming98}, where the authors utilized a morphological top-hat operator followed by a radial search procedure.
\section{Segmentation}
\label{sec_segm}
Segmentation refers to the partitioning of an image into disjoint regions that are homogeneous with respect to a chosen property such as luminance, color, texture, etc.~\cite{Sonka07}. Segmentation methods can be roughly classified into the following categories:
\begin{itemize}
 \item Histogram thresholding: These methods involve the determination of one or more histogram threshold values that separate the objects from the background.
 \item Clustering: These methods involve the partitioning of a color (feature) space into homogeneous regions using unsupervised clustering algorithms.
 \item Edge-based: These methods involve the detection of edges between the regions using edge operators.
 \item Region-based: These methods involve the grouping of pixels into homogeneous regions using region merging, region splitting, or both.
 \item Morphological: These methods involve the detection of object contours from predetermined seeds using the watershed transform.
 \item Model-based: These methods involve the modeling of images as random fields whose parameters are determined using various optimization procedures.
 \item Active contours (snakes and their variants): These methods involve the detection of object contours using curve evolution techniques.
 \item Soft computing: These methods involve the classification of pixels using soft-computing techniques including neural networks, fuzzy logic, and evolutionary computation.
\end{itemize}

Several issues should be considered when choosing a segmentation method:
\begin{itemize}
 \item Scalar vs.\ vector processing: Most segmentation methods are designed for scalar images. Although numerous vector image segmentation methods have been developed in the past decade, their usage is hindered by various factors including excessive computational time requirements and the difficulty of choosing an appropriate color space.
 \item Automatic vs.\ semi-automatic: Some segmentation methods are completely automated, whereas others require human interaction. For example, active contour methods often require the manual delineation of the initial contour, whereas seeded region growing methods require the specification of the initial region seeds.
 \item Number of parameters: Most segmentation methods have several parameters whose values need to be determined a priori. In general, the more the number of parameters, the harder the model selection (determination of the optimal parameters).
\end{itemize}

\section{Postprocessing}
\label{sec_post}
The result of the segmentation procedure is either a label image or a binary edge map. In order to obtain the lesion border, the segmentation output should be postprocessed. The precise sequence of postprocessing operations depends on the particular choice of the segmentation method. However, certain operations seem to be useful in general. These include:
\begin{itemize}
 \item Region merging: Ideally, the segmentation procedure is expected to produce two regions: the lesion and the background skin. However, since these regions are rarely homogeneous, segmentation methods often partition them into multiple subregions. In order to obtain a single lesion object, subregions that are part of the lesion should first be identified and then merged. This can accomplished in several ways:
  \begin{itemize}
   \item If the black frame of the image has already been removed, the background skin color can be estimated from the corners of the image and the subregions with similar color to the background skin can be eliminated, leaving only those subregions that are part of the lesion~\cite{Celebi08a,Celebi07b,Melli06}.
   \item Various color and texture features can be extracted from each region and a classifier can be trained to determine which features effectively discriminate between the regions that are part of the lesion and those that are part of the background skin~\cite{Fleming98}.
   \item The partitioning that maximizes the normalized texture gradient at the border and the total pairwise similarity among the regions within the lesion and those within the background skin can be determined by minimizing a cost function~\cite{Zhou08}.
  \end{itemize}
 \item Island removal: Islands (small isolated regions) in the label image can be eliminated using a binary area opening filter.
 \item Border smoothing: Most segmentation methods produce regions with ragged borders. More natural borders can be obtained by a variety of operations including majority filtering~\cite{Celebi08a}, morphological filtering~\cite{Schmid99a,Delgado08}, and curve fitting.
 \item Border expansion: In several studies, it was observed that the computer-detected borders were mostly contained within the dermatologist-determined borders. This is because the automated segmentation methods tend to find the sharpest pigment change, whereas the dermatologists choose the outmost detectable pigment. The discrepancy between the two borders can be reduced by expanding the computer-detected border using morphological filtering~\cite{Celebi08a}, Euclidean distance transform~\cite{Celebi08a}, or iterative region growing~\cite{Iyatomi06}.
\end{itemize}

\section{Evaluation}
\label{sec_eval}
Evaluation of the results seems to be one of the least explored aspects of the border detection task. As in the case of the more general image segmentation problem, there are two major evaluation methods: subjective and objective. The former involves the visual assessment of the border detection results by one or more dermatologists. Since there is no objective measure of quality involved, this technique does not permit parameter tuning or comparisons among automated border detection methods. On the other hand, objective evaluation involves the quantification of the border detection errors using dermatologist-determined borders. In the rest of this discussion, we refer to the computer-detected borders as \emph{automatic borders} and those determined by dermatologists as \emph{manual borders}. Most of the quantitative error metrics are based on the concepts of true/false positive/negative given in Table~\ref{tab_def} (here actual and detected pixels refer to a pixel in the ground-truth image and the corresponding pixel in the border detection output, respectively). These include:
\begin{itemize}
 \item XOR measure~\cite{Hance96} = $\frac{\mbox{FP} + \mbox{FN}}{\mbox{TP} + \mbox{FN}} \times 100\%$
 \item Sensitivity = $\frac{\mbox{TP}}{\mbox{TP} + \mbox{FN}} \times 100\%$ and Specificity = $\frac{\mbox{TN}}{\mbox{FP} + \mbox{TN}} \times 100\%$
 \item Precision (Positive Predictive Value) = $\frac{\mbox{TP}}{\mbox{TP} + \mbox{FP}} \times 100\%$ and Recall = Sensitivity
 \item Error probability =  $\frac{\mbox{FP} + \mbox{FN}}{\mbox{TP} + \mbox{FN} + \mbox{FP} + \mbox{TN}} \times 100\%$
\end{itemize}

\begin{table}[!ht]
\centering
\caption{ \label{tab_def} Definitions of true/false positive/negative }
\begin{tabular}{|ll|l|l|}
\hline
 \multicolumn{2}{|c|}{} & \multicolumn{2}{c|}{Detected Pixel} \\
\cline{3-4}
 &  & Lesion & Background \\
\hline
\multicolumn{1}{|l|}{Actual} & Lesion & True Pos. (TP) & False Neg. (FN) \\
\cline{2-4}
\multicolumn{1}{|l|}{Pixel} & Background & False Pos. (FP) & True Neg. (TN) \\
\hline
\end{tabular}
\end{table}

In a comprehensive study, Guillod \emph{et al.}~\cite{Guillod02} demonstrated that a single dermatologist, even one who is experienced in dermoscopy, cannot be used as an absolute reference for evaluating border detection accuracy. In addition, they emphasized that manual borders are not precise, with inter-dermatologist borders and even borders determined by the same dermatologist at different times showing significant disagreement, so that a probabilistic model of the border is preferred to an absolute gold-standard model. Accordingly, they used fifteen sets of borders drawn by five dermatologists over a minimum period of one month. A probability image for each lesion was constructed by associating a misclassification probability $p(i,j) = 1 - \frac{n(i,j)}{N}$ with each pixel (N: number of observations, $n(i,j)$: number of times pixel $(i,j)$ was selected as part of the lesion). For each automatic border $B$, the detection error was calculated as the mean probability of misclassification over the pixels inside the border, i.e. $\frac{\sum\limits_{(i,j) \in B} {p(i,j)}} {\mbox{TP} + \mbox{FP}} \times 100\%$.
\par
Iyatomi \emph{et al.}~\cite{Iyatomi06} modified Guillod \emph{et al.}'s approach by combining multiple manual borders that correspond to each lesion into one using the majority vote rule. The automatic borders were then compared against these combined ground-truth images. Celebi \emph{et al.}~\cite{Celebi08a} compared each automatic border against multiple manual borders independently.
\par
Unfortunately, the above-mentioned methods do not accurately capture the variations in the manual borders. For example, according to Guillod \emph{et al.}'s measure an automated border that is entirely inside the manual borders would get a very low error. Iyatomi \emph{et al.}'s method discounts the variation in the manual borders by simple majority voting, while Celebi \emph{et al.}'s approach does not produce a scalar error value, which makes comparisons more difficult.
\par
Recently, Celebi \emph{et al.}~\cite{Celebi09} proposed the use of an objective measure, the Normalized Probabilistic Rand Index~\cite{Unnikrishnan07}, which takes into account the variations in the manual borders. They demonstrated that the differences between four of the evaluated border detection methods were in fact smaller than those predicted by the commonly used XOR measure. Since the formulation of this measure is involved, the interested reader is referred to~\cite{Celebi09} and~\cite{Unnikrishnan07}.
\par
None of the above measures quantify the effect of border detection error upon the accuracy of the classifier. Loss of classification accuracy due to automatic border error can be measured as the difference of the classification accuracy using the manual borders and that using the automatic borders.
\section{Comparisons and Discussion}
\label{sec_comp}
Table~\ref{tab_methods} compares some of the recent border detection methods based on their operation mode (automatic vs.\ semi-automatic), color space, preprocessing steps, and segmentation method. Note that only those methods that are adequately described in the literature are included and the postprocessing steps are omitted since they are often not reported. The following observations are in order:
\begin{itemize}
 \item 16/18 methods are automated, reflecting the need for an unsupervised decision support system for clinical use.
 \item 11/18 methods operate on multiple color channels.
 \item 12/18 methods use a smoothing filter (the hair removal method described in~\cite{Lee97} also employs image smoothing).
 \item Clustering is the most popular segmentation method, which is probably due to the availability of robust algorithms.
\end{itemize}
Table~\ref{tab_eval} compares the border detection methods based on their evaluation methodology: the number of human experts who determined the manual borders, the number of images used in the evaluations (and the diagnostic distribution of these images if available), the number of automated methods used in the comparisons, and the measure used to quantify the border detection error. It can be seen that:
\begin{itemize}
 \item 9/18 studies rely on borders determined by a single dermatologist.
 \item Only 5/18 studies report the diagnostic distribution of their test images. This information is valuable given that not every diagnostic class is equally challenging from a border detection perspective. For example, it is often more difficult to detect the borders of melanomas and dysplastic nevi due to their irregular and fuzzy (hazy) border structure.
 \item 8/18 studies do not compare their results to those of any other automated method. This is partly due to the unavailability of public border detection software, as well as the non-existence of a public dermoscopy image database.
  \item Recent studies used objective measures to determine the validity of their results, whereas earlier studies relied on visual assessment. XOR measure is the most popular objective error function despite the fact that it is not trivial to extend this measure to capture the variations in multiple manual borders.
\end{itemize}

We believe that in a systematic border detection study:
\begin{enumerate}
 \item The image acquisition procedure should be described in sufficient detail.
 \item The test image set should be selected randomly from a large and diverse image database.
 \item The test image set should be large enough to ensure statistically valid conclusions.
 \item The diagnostic distribution of the test image set should be stated.
 \item Algorithms with reasonable computational requirements should be used.
 \item The results should be evaluated using borders determined by multiple dermatologists.
 \item The results should be compared to those of published border detection methods.
 \item The border detection procedure should be described in sufficient detail.
\end{enumerate}
Note that all of the abovementioned criteria except for (5) and (8) can be satisfied by using a public dermoscopy image set. Therefore, the creation of such a benchmark database should be prioritized in order to improve the quality of future border detection studies.
\par
We must recognize the limitations of automated systems at present in comparison to experienced dermoscopy practitioners, who have the unparalleled ability to correctly identify borders in several situations. First, 'collision tumors', contiguity of lesions of more than one type, are fairly common. The most common of these collisions and the situation most likely to cause error in automatic diagnosis is the collision between a malignancy and a benign lesion, often a lentigo, which is common in sun-damaged skin. Dermatologists are better able to define what is normal for a given patient's background skin, even when the background skin includes lentigines that can falsely enlarge the border. Second, some significant melanoma features may be lost without higher level knowledge that enables the inclusion of such features within the lesion. Scar-like regression, an area of pallor that is a significant feature for melanoma in-situ, is excluded from the lesion if it is close to the border in all border detection methods that we have examined. Third, dermatologists vary the borders according to the diagnosis. Dermatologist borders include the halo in halo nevi, and the pale rim of basal cell carcinomas, but exclude the surrounding reactive erythema of irritated lesions. It is likely that some higher-level knowledge will need to be included in automated border detection methods to accomplish the ultimate purpose of these computer systems: to achieve higher diagnostic accuracy.
\begin{table}
\centering
\caption{ \label{tab_methods} Characteristics of border detection methods (nr: not reported, KL\{$C$\}: KL transform of the $C$ color space)}
\begin{tabular}{|c|c|c|c|c|c|}
\hline
Ref.\ & Year & Mode & Color Space (\# Channels) & Preprocessing Steps & Segmentation Method\\
\hline
\hline
\cite{Celebi08a} & 2008 & Auto & RGB (3) & Marginal MF, Black frame removal & Region-Based\\
\cite{Zhou08} & 2008 & Auto & $L^*a^*b^*$ (3) & nr & Clustering\\
\cite{Delgado08} & 2008 & Auto & $\mbox{IHP\{RGB\}}$ (1) & Contrast enhancement, Hair removal~\cite{Lee97} & Clustering\\
\cite{Mendonca07} & 2007 & Semi & Luminance (1) & GF & Active Contours\\
\cite{Mendonca07} & 2007 & Auto & RGB (3) & nr & Active Contours\\
\cite{Celebi07b} & 2007 & Auto & RGB (3) & Marginal MF, Approx.\ Lesion Localization & Region-Based\\
\cite{Iyatomi06} & 2006 & Auto & B / RGB (1) & GF & Thresholding\\
\cite{Melli06} & 2006 & Auto & RGB (3) & nr & Clustering\\
\cite{Erkol05} & 2005 & Auto & Luminance (1) & GF, Auto.\ Snake Initialization & Active Contours\\
\cite{Galda03} & 2003 & Auto & $L^*u^*v^*$ (3) & Marginal MF & Clustering\\
\cite{Cucchiara02} & 2002 & Auto & $\mbox{KL} \{ L^*a^*b^* \}$ (2) & GF & Clustering\\
\cite{Hintz-Madsen01} & 2001 & Auto & $\mbox{KL\{RGB\}}$ (1) & Marginal MF & Thresholding\\
\cite{Haeghen00} & 2000 & Semi & $L^*a^*b^* $ (3) & nr & Active Contours\\
\cite{Donadey00} & 2000 & Auto & I / HSI (1) & MF & Soft Computing\\
\cite{Schmid99a} & 1999 & Auto & $\mbox{KL} \{ L^*u^*v^* \} $ (2) & MF & Clustering\\
\cite{Schmid99b} & 1999 & Auto & $ L^*a^*b^* $ (3) & Vector ADF, Hair removal & Morphological\\
\cite{Gao98} & 1998 & Auto & RGB (3) & nr & Region-Based\\
\cite{Gao98} & 1998 & Auto & $\mbox{KL\{RGB\}} (1) $ & nr & Model-Based\\
\hline
\end{tabular}
\end{table}

\begin{table}
\centering
\caption{ \label{tab_eval} Evaluation of border detection methods (b: benign, m: melanoma) }
\begin{tabular}{|c|c|c|c|c|}
\hline
Ref.\ & \# Experts & \# Images (Distr.) & \# Comp.\ & Error Measure\ (Value) \\
\hline
\hline
\cite{Celebi08a} & 3 & 90 (65 b / 25 m) & 4 & XOR (10.63\%)\\
\cite{Zhou08} & 1 & 67 & 0 & XOR (14.63\%) \\
\cite{Delgado08} & 1 & 100 (70 b / 30 m) & 3 & XOR (2.73\%)\\
\cite{Mendonca07} & 1 & 50 & 2 & Error prob.\ (16\%)\\
\cite{Mendonca07} & 1 & 50 & 2 & Error prob.\ (21\%)\\
\cite{Celebi07b} & 2 & 100 (70 b / 30 m) & 3 & XOR (12.02\%)\\
\cite{Iyatomi06} & 5 & 319 (244 b / 75 m) & 1 & Prec.\ (94.1\%) \& Rec.\ (95.2\%)\\
\cite{Melli06} & nr & 117 & 3 & Sens.\ (95\%) \& Spec.\ (96\%)\\
\cite{Erkol05} & 2 & 100 (70 b / 30 m) & 1 & XOR (15.59\%)\\
\cite{Galda03} & 0 & nr & 0 & nr\\
\cite{Cucchiara02} & 0 & 600 & 0 & Visual\\
\cite{Hintz-Madsen01} & 0 & nr & 0 & nr\\
\cite{Haeghen00} & 5 & 30 & 0 & Visual\\
\cite{Donadey00} & 1 & 30 & 0 & Visual\\
\cite{Schmid99a} & 1 & 400 &0 & Visual\\
\cite{Schmid99b} & 1 & 300  & 0 & Visual\\
\cite{Gao98} & 1 & 57 & 5 & XOR (36.50\%)\\
\cite{Gao98} & 1 & 57 & 5 & XOR (24.71\%)\\
\hline
\end{tabular}
\end{table}

\section*{Acknowledgments}
This publication was made possible by grants from The Louisiana Board of Regents (LEQSF2008-11-RD-A-12) and The National Institutes of Health (SBIR \#2R44 CA-101639-02A2). Its contents are solely the responsibility of the authors and do not necessarily represent the official views of The Louisiana Board of Regents or the NIH.


\begin{thebibliography}{99}

\begin{singlespace}

\bibitem{Jemal07} Jemal A, Siegel R, Ward E \emph{et al.} Cancer Statistics, 2008. CA: A Cancer Journal for Clinicians 2008; 58(2): 71--96, 2008.

\bibitem{Argenziano02} Argenziano G, Soyer HP, De Giorgi V \emph{et al.} Dermoscopy: A Tutorial. Milan, Italy: EDRA Medical Publishing \& New Media, 2002.

\bibitem{Menzies03} Menzies SW, Crotty KA, Ingwar C, McCarthy WH. An Atlas of Surface Microscopy of Pigmented Skin Lesions: Dermoscopy. Sydney, Australia: McGraw-Hill, 2003.

\bibitem{Steiner93} Steiner K, Binder M, Schemper M \emph{et al.} Statistical Evaluation of Epiluminescence Dermoscopy Criteria for Melanocytic Pigmented Lesions. Journal of American Academy of Dermatology 1993; 29(4): 581--588.

\bibitem{Binder95} Binder M, Schwarz M, Winkler A \emph{et al.} Epiluminescence Microscopy. A Useful Tool for the Diagnosis of Pigmented Skin Lesions for Formally Trained Dermatologists. Archives of Dermatology 1995; 131(3): 286--291.

\bibitem{Fleming98} Fleming MG, Steger C, Zhang J \emph{et al.} Techniques for a Structural Analysis of Dermatoscopic Imagery. Computerized Medical Imaging and Graphics 1998; 22(5): 375--389.

\bibitem{Celebi07a} Celebi ME, Kingravi HA, Uddin B \emph{et al.} A Methodological Approach to the Classification of Dermoscopy Images. Computerized Medical Imaging and Graphics 2007; 31(6): 362--373.

\bibitem{Pratt07} Pratt WK. Digital Image Processing: PIKS Inside. Hoboken, NJ: John Wiley \& Sons, 2007.

\bibitem{Delgado08} Delgado D, Butakoff C, Ersboll BK, Stoecker WV. Independent Histogram Pursuit for Segmentation of Skin Lesions. IEEE Trans.\ on Biomedical Engineering 2008; 55(1): 157--161.

\bibitem{Celebi07c} Celebi ME, Kingravi HA, Aslandogan YA. Nonlinear Vector Filtering for Impulsive Noise Removal from Color Images. Journal of Electronic Imaging 2007; 16(3): 033008 (21 pages).

\bibitem{Schmid99a} Schmid P. Segmentation of Digitized Dermatoscopic Images by Two-Dimensional Color Clustering. IEEE Trans.\ on Medical Imaging 1999; 18(2): 164--171.

\bibitem{Celebi08a} Celebi ME, Kingravi HA, Iyatomi H \emph{et al.} Border Detection in Dermoscopy Images Using Statistical Region Merging. Skin Research and Technology 2008; 14(3): 347--353.

\bibitem{Geusebroek03} Geusebroek J-M, Smeulders AWM, van de Weijer J. Fast Anisotropic Gauss Filtering. IEEE Trans.\ on Image Processing 2003; 12(8): 938--943.

\bibitem{Perreault07} Perreault S, H\'{e}bert P. Median Filtering in Constant Time. IEEE Trans.\ on Image Processing 2007, 16(9): 2389--2394.

\bibitem{Lee97} Lee TK, Ng V, Gallagher R. \emph{et al.} Dullrazor: A Software Approach to Hair Removal from Images. Computer in Biology and Medicine 1997; 27(6): 533--543.

\bibitem{Zhou08} Zhou H, Chen M, Gass R \emph{et al.} Feature-Preserving Artifact Removal from Dermoscopy Images. Proc. of the SPIE Medical Imaging 2008 Conf.\, 6914: 69141B--69141B-9.

\bibitem{Wighton08} Wighton P, Lee TK, Atkins MS Dermascopic Hair Disocclusion Using Inpainting. Proc. of the SPIE Medical Imaging 2008 Conf.\, 6914: 691427--691427-8.

\bibitem{Sonka07} Sonka M, Hlavac V, Boyle R. Image Processing, Analysis, and Machine Vision. Cengage-Engineering, 2007.

\bibitem{Celebi07b} Celebi ME, Aslandogan YA, Stoecker WV \emph{et al.} Unsupervised Border Detection in Dermoscopy Images. Skin Research and Technology 2007; 13(4): 454--462.

\bibitem{Melli06} Melli R, Grana C, Cucchiara R. Comparison of Color Clustering Algorithms for Segmentation of Dermatological Images. Proc.\ of the SPIE Medical Imaging 2006 Conf.\, pp. 3S1--9.

\bibitem{Iyatomi06} Iyatomi H, Oka H, Saito M \emph{et al.} Quantitative Assessment of Tumor Extraction from Dermoscopy Images and Evaluation of Computer-based Extraction Methods for Automatic Melanoma Diagnostic System. Melanoma Research 2006; 16(2): 183--190.

\bibitem{Hance96} Hance GA, Umbaugh SE, Moss RH, Stoecker WV. Unsupervised Color Image Segmentation with Application to Skin Tumor Borders. IEEE Engineering in Medicine and Biology 1996; 15(1): 104--111.

\bibitem{Guillod02} Guillod J, Schmid-Saugeon P, Guggisberg D \emph{et al.} Validation of Segmentation Techniques for Digital Dermoscopy. Skin Research and Technology 2002; 8(4): 240--249.
  
\bibitem{Celebi09} Celebi ME, Iyatomi H, Schaefer G \emph{et al.} An Improved Objective Evaluation Measure for Border Detection in Dermoscopy Images. Skin Research and Technology 2009; 15(4): 444--450.

\bibitem{Unnikrishnan07} Unnikrishnan R, Pantofaru C, Hebert M. Toward Objective Evaluation of Image Segmentation Algorithms. IEEE Trans.\ on Pattern Analysis and Machine Intelligence 2007; 29(6): 929--944.

\bibitem{Mendonca07} Mendonca T, Marcal ARS, Vieira A \emph{et al.} Comparison of Segmentation Methods for Automatic Diagnosis of Dermoscopy Images. Proc.\ of the 29th IEEE EMBS Annual Int.\ Conf.\, 1: 6572--6575.

\bibitem{Erkol05} Erkol B, Moss RH, Stanley RJ, Stoecker WV, Hvatum E. Automatic Lesion Boundary Detection in Dermoscopy Images Using Gradient Vector Flow Snakes. Skin Research and Technology 2005; 11(1): 17--26.

\bibitem{Galda03} Galda H, Murao H, Tamaki H, Kitamura S. Skin Image Segmentation Using a Self-Organizing Map and Genetic Algorithms. Trans.\ of the Institute of Electrical Engineers of Japan - Part C 2003; 123(11): 2056--2062.

\bibitem{Cucchiara02} Cucchiara R, Grana C, Seidenari S, Pellacani G. Exploiting Color and Topological Features for Region Segmentation with Recursive Fuzzy c-means. Machine Graphics and Vision 2002; 11(2/3): 169--182.

\bibitem{Hintz-Madsen01} Hintz-Madsen M, Hansen LK, Larsen J, Drzewiecki K. A Probabilistic Neural Network Framework for the Detection of Malignant Melanoma. Artificial Neural Networks in Cancer Diagnosis, Prognosis and Patient Management, pp. 141--183, 2001.

\bibitem{Haeghen00} Haeghen YV, Naeyaert JM, Lemahieu I. Development of a Dermatological Workstation: Preliminary Results on Lesion Segmentation in  CIE $L^*A^*B^*$ Color Space. Proc. of the Int. Conf.\ on Color in Graphics and Image Processing.

\bibitem{Donadey00} Donadey T, Serruys C, Giron A \emph{et al.} Boundary Detection of Black Skin Tumors Using an Adaptive Radial-based Approach. Proc.\ of the SPIE Medical Imaging 2000 Conf.\, 3379: 810--816.

\bibitem{Schmid99b} Schmid P. Lesion Detection in Dermatoscopic Images Using Anisotropic Diffusion and Morphological Flooding. Proc.\ of the IEEE ICIP 1999 Conf.\, 3: 449--453.

\bibitem{Gao98} Gao J, Zhang J, Fleming MG \emph{et al.} Segmentation of Dermatoscopic Images by Stabilized Inverse Diffusion Equations. Proc.\ of the IEEE ICIP 1998 Conf., 3: 823--827.

\end{singlespace}
\end{thebibliography}
\end{document}